\documentclass[letterpaper, 10 pt, conference]{ieeeconf}  

\IEEEoverridecommandlockouts                              

\usepackage{mathbbol}
\usepackage{algorithm}
\usepackage{xcolor}
\usepackage{algorithmic}
\usepackage{hyperref}
\usepackage{euscript}[mathcal]
\usepackage{multirow}
\usepackage{cite}
\usepackage{amsmath}
\usepackage{graphicx}
\usepackage[caption=true]{subfig}
\usepackage{algorithm}
\usepackage{algorithmic}
\usepackage{xcolor}
\usepackage{graphics} 
\usepackage{amsfonts}
\usepackage{multirow}

\newtheorem{remark}{Remark}
\newcommand{\x}{\mathbf{x}}
\title{\LARGE \bf
A Supervised Feature Selection Method For Mixed-Type Data using Density-based Feature Clustering
}
\author{Xuyang Yan$^{1}$, Mrinmoy Sarkar$^{1}$,
Biniam Gebru$^{1}$,
Shabnam Nazmi$^{1}$,
and Abdollah Homaifar$^{1,*}$ 
\thanks{$^{1}$North Carolina A\&T State University, Greensboro, North Carolina, US, 27411. {\tt Emails: xyan@aggies.ncat.edu;
msarkar@aggies.ncat.edu; btgebru@aggies.ncat.edu; snazmi@aggies.ncat.edu;
homaifar@ncat.edu}}%
\thanks{$^{*}$ Corresponding Author}%
}

\begin{document}

\maketitle
\thispagestyle{empty}
\pagestyle{empty}

\begin{abstract}
Feature selection methods are widely used to address the high computational overheads and \textit{curse of dimensionality} in classifying high-dimensional data. Most conventional feature selection methods focus on handling homogeneous features, while real-world datasets usually have a mixture of continuous and discrete features. Some recent mixed-type feature selection studies only select features with high relevance to class labels and ignore the redundancy among features. The determination of an appropriate feature subset is also a challenge. In this paper, a supervised feature selection method using density-based feature clustering (SFSDFC) is proposed to obtain an appropriate final feature subset for mixed-type data. SFSDFC decomposes the feature space into a set of disjoint feature clusters using a novel density-based clustering method. Then, an effective feature selection strategy is employed to obtain a subset of important features with minimal redundancy from those feature clusters. Extensive experiments as well as comparison studies with five state-of-the-art methods are conducted on SFSDFC using thirteen real-world benchmark datasets and results justify the efficacy of the SFSDFC method. 


\end{abstract}
\section{INTRODUCTION}
The high computational overheads and \textit{curse of dimensionality} \cite{duda2012pattern} inspire the ever-increasing research interests in feature selection (FS). Unlike feature extraction methods, FS aims to obtain a subset of relevant or non-redundant features from the original feature space without losing important information and data structure. The mixture of continuous and discrete variables is widely used as the primary representation of many real-world high-dimensional datasets \cite{de2013analysis}. Hence, it is necessary to investigate solutions for heterogeneous feature selection problems.\\
\indent Conventional supervised feature selection methods are divided into three main categories \cite{dash1997feature,dong2018feature,yan2018unsupervised}: \textit{filter, wrapper}, and \textit{hybrid}. The \textit{filter} methods evaluate the importance of features using the intrinsic data properties such as correlation or dependency among attributes. \textit{Wrapper} methods utilize a statistical model to assess features based on their influence on the model performance. The \textit{hybrid} approaches integrate \textit{filter} and \textit{wrapper} methods to balance the trade-off between computational efficiency and performance. Under these three main groups, numerous supervised feature selection approaches are proposed to handle data with only continuous or discrete features \cite{peng2005feature,hall1999correlation,kononenko1994estimating,nie2010efficient,solorio2020supervised} while few attentions are employed on feature selection problems with mixed-type data. \\
\indent In the mixed-type feature selection problems, one popular strategy is to transform data from heterogeneous feature space into homogeneous feature space \cite{solorio2020supervised,barcelo2012geometrical,cohen2013applied}. This strategy usually suffers from the risk of information loss during the transformation procedure \cite{hu2008neighborhood}. Another well-known strategy performs feature selection on continuous or discrete features separately and then combines the selected features \cite{tang2007feature,hu2008neighborhood,liang2012determining,chen2013attribute}. In \cite{liang2012determining}, the authors defined two different feature evaluation criteria for continuous and discrete features independently. Then, a combination of feature selection results from these two different criteria is obtained as the final selected feature subset. \\
\indent In recent years, fuzzy rough set feature selection methods are extensively investigated \cite{hu2009selecting,hu2011measuring,wang2016efficient,zhang2016feature,kim2018rough}. In \cite{hu2009selecting}, the fuzzy rough set theory was employed to conduct attribute reduction for mixed data first and a number of extensions are introduced in \cite{hu2011measuring,wang2016efficient,zhang2016feature,kim2018rough}. These approaches integrate fuzzy rough set theory with information entropy to measure the dependency among heterogeneous feature types. To satisfy the monotonicity criteria for general fuzzy systems, a rough set based conditional entropy function is proposed in \cite{zhang2016feature}. Despite the effectiveness of fuzzy rough set based feature selection methods, the high computational complexity strongly limited their applications for large-scale datasets. Furthermore, most rough set based feature selection methods adopt a sequential feature selection procedure to obtain the final feature subset and a more effective strategy should be considered to determine the size of the selected feature subset.\\
\indent \textcolor{black}{Several recent \textit{filter}-based feature selection approaches attempted to reduce feature redundancy through feature clustering analysis \cite{cheung2012unsupervised,zhu2017subspace,zhu2018co,bandyopadhyay2014integration,yan2020efficient}. Features are separated into a set of clusters based on their similarity such that highly redundant features are grouped together. Then, a set of non-redundant features are selected from those feature clusters. These approaches have shown promising performance in reducing feature redundancy without losing important information. However, few of them considered continuous and discrete features simultaneously. In \cite{yan2020efficient}, although the authors employed a modified clustering procedure to perform continuous or discrete feature selection separately, it assumed that all features are homogeneous. Besides, the relevance of features with respect to the class label is not considered.}  \\
\indent Motivated by the challenges described above, we propose a \textit{filter}-based \textbf{S}upervised \textbf{F}eature \textbf{S}election framework through a \textbf{D}ensity-based \textbf{F}eature \textbf{C}lustering procedure, namely SFSDFC, to address the mixture of continuous and discrete features. First of all, a practical heuristic is used to systematically separate the continuous and discrete features. Then, we perform the feature cluster analysis and subsequent feature selection on continuous and discrete features independently. We develop a generalized density-based clustering procedure that is applicable to both the continuous and discrete features. Also, a simple yet effective feature selection strategy is introduced to obtain a subset of relevant and representative features using the label information. Finally, we combine the selected continuous and discrete feature subset together as the final feature subset. 
Overall, the primary contributions of this paper are summarized below.
\begin{itemize}
    \item Developed a novel density-based feature clustering technique to partition the feature space into an appropriate number of feature clusters without prior knowledge.
    \item Introduced a simple yet effective feature selection strategy to choose both relevant and representative features from feature clusters. 
    \item Conducted extensive experiments and comprehensive comparison studies to justify the efficacy of the SFSDFC method. 
\end{itemize}

The remainder of this paper is organized as follows: Section \ref{section:2} defines the preliminaries of this research, including basic notations, feature similarity measures, and feature relevance metrics. The details of SFSDFC are described in Section \ref{section:3}, and the computational complexity of SFSDFC is analyzed in Section \ref{section:4}. Section \ref{section:5} presents the experimental results and comparison study between SFSFC and the state-of-the-art methods. Finally, concluding remarks and future works are outlined in Section \ref{section:6}.
\section{PRELIMINARIES} \label{section:2}
\indent This section discusses basic notations, feature similarity measures, and feature relevance metrics, respectively.
\subsection{Basic notations}
Let an labeled dataset be $D$ and the feature space be $F$ such that $D=\{(\x_i, y_i)|\x_i\in F, y_i\in Y, i= 1,\ldots n\}$  where $F=\{f_{1},f_{2},...,f_{m}\}$. The notations $n$ and $m$ refer to the number of samples and features, respectively. $F_{cont}$ and $F_{disc}$ represent the continuous and discrete feature subset, respectively. Moreover, we use $FC_{cont}$ and $FC_{disc}$ to denote the obtained feature clusters in $F_{cont}$ and $F_{disc}$.  The final feature subset is denoted as $F_{s}$. 
\subsection{Feature similarity measures}
The Maximal Information Compression Index (MICI) \cite{mitra2002unsupervised} and Symmetrical Uncertainty (SU) \cite{press1988numerical} are used to measure the similarity among the continuous and discrete features, respectively. The definitions of these two metrics are discussed below. 
\subsubsection{MICI} It measures the similarity among continuous features based on the linear dependency relationship. With a two-by-two covariance matrix constructed for two continuous features, MICI is defined as follows:
\normalsize
\small
\begin{multline}
    MICI = Var(f_{j}) + Var(f_{k}) - \\\sqrt{\left(Var(f_{j})+Var(f_{k})\right)^2 - 4Var(f_{j})Var(f_{k})(1-\rho(f_{j},f_{k})^2)}. \label{eq:2}
\end{multline}
\normalsize
\indent Here, $Var(f_{j})$ and $Var(f_{k})$ denote the variance of the $j^{th}$ and the $k^{th}$ feature, respectively. The notation $\rho(f_{j},f_{k})$ is the Pearson correlation coefficient between the features $j$ and $k$. The value of \textit{MICI} ranges from zero to one, where $0$ indicates that features $j$ is completely dependent on feature $k$ while $1$ implies zero linear dependency.   
\subsubsection{SU} It is derived from the mutual information \cite{kraskov2004estimating} between the probability distributions of two discrete features. Equation \ref{eq:3} explains the calculation of \textit{SU} as follows.
\begin{equation}
SU = \frac{2\times Gain(f_{j}|f_{k})}{H(f_{j}) + H(f_{k})},\label{eq:3}
\end{equation}
\normalsize
where $H(f_{j})$ and $H(f_{k})$ refer to the entropy of the $j^{th}$ and $k^{th}$ features, respectively. $Gain(f_{j}|f_{k})$ represents the information gain of $f_{j}$ provided by $f_{k}$.
\subsection{Feature relevance metric}
Mutual Information (MI) \cite{kraskov2004estimating} is used to measure the amount of information provided by a single feature with respect to class labels. Assuming $I(Y;f_{j})$ to be the MI between feature $j$ and class label space $Y$, the definition of MI is expressed as follows.
\begin{equation}
    I(Y;f_{j})=H(Y) -H(Y|f_{j}). \label{eq:4}
\end{equation}
The terms $H(Y)$ refers to the entropy of the label space and $H(Y|f_{j})$ denotes the conditional entropy of $Y$ provided by feature $j$. 
\section{PROPOSED METHODOLOGY} \label{section:3}
In this section, the proposed SFSDFC framework is discussed in terms of its three main steps: (i) separating continuous and discrete features; (ii) feature cluster analysis; and (iii) subsequent feature selection strategy. 
\subsection{Separation between continuous and discrete features}
\indent \textcolor{black}{As mentioned previously, SFSDFC separates $F$ into two subsets: (i) continuous subset ($F_{cont}$) and discrete subset ($F_{disc}$). For datasets with known variable types, we simply use this prior information to obtain the continuous and discrete feature subsets. For datasets with unknown feature types, SFSDFC employs a popular heuristic approach by comparing the number of unique values for a single feature with a threshold variable $\epsilon$. A feature is considered to be discrete when the number of its unique values falls below $\epsilon$; otherwise, it is considered as a continuous feature. Motivated by the rule of thumb principle for choosing the value of $k$ in $k$-nearest neighbor ($k$NN) rule \cite{altman1992introduction}, the value of $\epsilon$ is set to $\sqrt{n}$ in this paper. This procedure is repeated for each feature. Finally, two feature subsets are obtained for the subsequent feature cluster analysis as well as the feature selection procedure. }  
\subsection{Density-based feature clustering analysis} 
A recently developed density-based clustering approach, namely FPS-clustering \cite{yan2017novel}, is used to explore the cluster structure of data without any prior knowledge or parameter optimization. In \cite{yan2020efficient}, the authors extended FPS-clustering to feature clustering analysis for completely continuous or discrete features by proposing two different cluster merge schemes. Hence, we develop a new feature clustering approach by integrating the FPS-clustering with a universal cluster merge strategy, which can be applied to both continuous and discrete features. We utilize the potential cluster center search procedure from \cite{yan2020efficient} to discover a set of potential centers and form temporary potential clusters. Then, we perform the merge among those temporary clusters based on the cluster radius rather than density values. To search for temporary potential clusters, a Gaussian kernel density function is used to approximate the density distribution and it is expressed as follows.
\begin{equation}
    P(f_{j}) = \sum_{k=1}^{m}{(e^{\frac{Dist(f_{j},f_{k})}{\beta}})^{\gamma}}. \label{eq:6}
\end{equation}
Here, $P(f_{j})$ denotes the density value of the $j^{th}$ feature and $Dist(f_{j},f_{k})$ is the dissimilarity between $f_{j}$ and $f_{k}$. The terms $\beta$ refers to the normalization parameter and $\gamma$ is the stabilization parameter. We adopt the same parameter estimation procedure from \cite{yan2017novel} to obtain the values of $\beta$ and $\gamma$. Algorithm \ref{algo2} summarizes the density-based feature clustering procedure.\\
\begin{algorithm}[tb]
\footnotesize
\caption{\textcolor{black}{Feature cluster analysis}}
\label{algo2}
\textbf{Input}: $F_{cont},F_{disc}$\\
\textbf{Parameter}: A set of temporary continuous cluster centers $FC_{tp}^{c}$, a set of temporary discrete cluster centers $FC_{tp}^{d}$, a set of continuous feature clusters $FC_{cont}$, a set of discrete feature clusters $FC_{disc}$.\\
\textbf{Output}: $FC_{cont}$, $FC_{disc}$
\begin{algorithmic}[1] 
\STATE Calculate the density values for $F_{cont}$ and $F_{disc}$ using equation \ref{eq:6}
\STATE Search for $FC_{tp}^{c}$ and $FC_{tp}^{d}$ in $F_{cont}$ and $F_{disc}$ respectively
\STATE Perform the merge in $FC_{tp}^{c}$ and $FC_{tp}^{d}$ based on remark \ref{remark1}
\STATE Obtain $FC_{cont}$ and $FC_{disc}$ from the merging results
\STATE \textbf{return} $FC_{cont}$, $FC_{disc}$
\end{algorithmic}
\end{algorithm}
\indent Algorithm \ref{algo2} starts by estimating the density values of features in the $F_{cont}$ and $F_{disc}$ using equation \ref{eq:6}. With the estimated density values, we conduct the search of temporary potential feature cluster centers in these two subsets using the fitness proportionate sharing (FPS) strategy \footnote{The details of the FPS strategy can found be in \cite{yan2017novel}.}. For $F_{cont}$, we rank all features into descending order with respect to their density values and select the highest ranked feature as the temporary center. Then, we scale down the density values of features that fall inside the neighborhood of the current temporary center. This helps to reduce the chance for those features to become the next temporary center and thus avoid unnecessary explorations for potential cluster centers. The same procedure is applied on $F_{disc}$ to obtain a set of temporary potential clusters. After obtaining temporary potential clusters, a cluster merge procedure is implemented on both $F_{cont}$ and $F_{disc}$ to merge highly overlapped temporary potential clusters. Let two neighboring temporary feature clusters be $FC_{tp_{1}}$ and $FC_{tp_{2}}$, $r_{1}$ and $r_{2}$ are their radius, respectively. Also, let $d_{12}$ be the distance between the centers of $FC_{tp_{1}}$ and $FC_{tp_{2}}$. The following remark defines the merge criterion for highly-overlapped temporary feature clusters.

\begin{remark}
Two neighboring temporary feature clusters $FC_{tp_{1}}$ and $FC_{tp_{2}}$ should be merged if: 
\begin{equation}
    d_{12} < r_{1}+r_{2}. \label{eq:7}
\end{equation}
\label{remark1}
\end{remark}

\begin{figure}[h]
\centering
\includegraphics[width=2.0in]{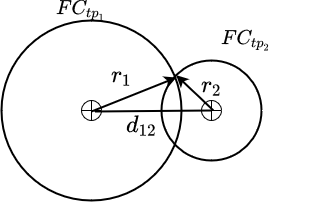}
\caption{An example of two overlapped clusters.}
\label{fig: example}
\end{figure}

This idea is motivated by the property of two overlapped circles and it has been successfully used to merge highly overlapped clusters in \cite{xiong2012similarity,qi2017effective}. 
An example of two overlapped clusters is shown in Fig. \ref{fig: example}, where $FC_{tp_{1}}$ and $FC_{tp_{2}}$ are overlapped and a triangle can be formed by $r_{1}$, $r_{2}$, and $d_{12}$. From this aspect, equation \ref{eq:7} can also be justified by the triangular principle.
\subsection{Selection of representative and relevant features}
The feature clustering procedure groups the features with high redundancy together and separates features that are less redundant into different clusters. Thus, each feature cluster center can be considered as the most representative feature for a set of redundant features. In \cite{yan2020efficient}, feature cluster centers are selected to form a set of non-redundant representative features. In this paper, we take advantage of the label information and propose a new feature selection criteria to obtain  a more reliable feature subset with minimal redundancy by simultaneously considering the feature relevance and redundancy. The details of the proposed feature selection criteria are described in Algorithm \ref{algo3}.\\
\begin{algorithm}[tb]
\footnotesize
\caption{\textcolor{black}{Relevant and representative feature selection}}
\label{algo3}
\textbf{Input}: $FC_{cont},FC_{disc}$\\
\textbf{Parameter}: The continuous feature cluster center $FC_{c_{1}}$, the discrete feature cluster center $FC_{c_{2}}$, a set of selected continuous features $F_{cont_{s}}$, a set of selected discrete features $F_{disc_{s}}$.\\
\textbf{Output}: $F_{s}$
\begin{algorithmic}[1] 
\STATE $F_{cont_{s}}=\emptyset$, $F_{disc_{s}}=\emptyset$
\FOR{$i=1$ to $|FC_{cont}|$}
\STATE Perform feature relevance evaluation on $FC^{i}_{cont}$ using equation \ref{eq:4}
\STATE Identify the most relevant feature as $FR^{i}_{cont}$
\IF{$FC^{i}_{c_{1}}$ is not the most relevant feature}
\STATE $F_{cont_{s}}=F_{cont_{s}}\bigcup \left(FC^{i}_{c_{1}} \bigcup FR^{i}_{cont}\right)$
\ELSE
\STATE $F_{cont_{s}}=F_{cont_{s}}\bigcup FC^{i}_{c_{1}}$
\ENDIF
\ENDFOR
\FOR{$i=1$ to $|FC_{disc}|$}
\STATE Perform feature relevance evaluation on features in $FC^{i}_{disc}$
\STATE Identify the most relevant feature as $FR^{i}_{disc}$
\IF{$FC^{i}_{c_{2}}$ is not the most relevant feature}
\STATE $F_{disc_{s}}=F_{disc_{s}}\bigcup\left(FC^{i}_{c_{2}}\bigcup FR^{i}_{disc}\right)$
\ELSE
\STATE $F_{disc_{s}}=F_{disc_{s}}\bigcup FC^{i}_{c_{2}}$
\ENDIF
\ENDFOR
\STATE $F_{s}=F_{cont_{s}}\bigcup F_{disc_{s}}$
\STATE \textbf{return} $F_{s}$
\end{algorithmic}
\end{algorithm}
\indent As described in algorithm \ref{algo3}, we perform the selection of features in continuous feature clusters and discrete feature clusters separately. For both the continuous and discrete feature clusters, the evaluation of feature relevance is performed within each feature cluster. Specifically, the feature cluster center is considered as the selected feature when it has the highest feature relevance. Otherwise, we select both the center and the most relevant feature. Finally, the combination of the selected continuous and discrete feature subsets is used as the final feature subset $F_{s}$. 
\section{COMPLEXITY ANALYSIS} \label{section:4}
Assume $D$ has $n$ samples and each sample has $m$ features, the time complexity of the SFSDFC framework is analyzed below:
\begin{itemize}
    \item In feature separation procedure, each feature takes $O(n)$ iterations to count the number of unique values and thus requires a total number of $O(mn)$ computations.
    \item For feature clustering procedure, the feature similarity evaluation requires $O(nm^{2})$ computations and the ranking of features takes $O(mlog(m))$ computations.
    \item During feature selection process, the feature relevance evaluation causes $O(mn)$ calculations. 
\end{itemize}

Therefore, the total time complexity of SFSDFC method is: $O(mn)+O(m^2n)+O(nlogn)+O(mn)\approx O(m^2n)$.

\section{EXPERIMENTS AND RESULTS} \label{section:5}
This section presents the experimental results and comparison studies of SFSDFC with five popular existing methods using thirteen well-known benchmark datasets. 
\subsection{Benchmark datasets}
Thirteen real-world benchmark datasets from the UCI machine learning repository \cite{Dua:2017} are used to evaluate the efficacy of the SFSDFC framework. Among these datasets, eleven of them have a mixture of continuous and discrete features. In order to verify the efficacy of SFSDFC on homogeneous features, the remaining two datasets only have continuous or discrete features. Table \ref{tabel:1} provides the details of all benchmark datasets in terms of sample size, feature size, number of continuous \& discrete features, and number of classes.
\begin{table}[h]
\caption{Dataset properties.} \label{tabel:1}
\resizebox{\columnwidth}{!}{
\begin{tabular}{l|l|c|c|c|c}
\hline
\multirow{2}{*}{Datasets} & \multirow{2}{*}{Samples} & \multicolumn{3}{c|}{Features}                                                                  & \multirow{2}{*}{Classes} \\ \cline{3-5}
                          &                          & \multicolumn{1}{l|}{Total} & \multicolumn{1}{l|}{Continuous} & \multicolumn{1}{l|}{Discrete} &                          \\ \cline{3-6} \hline
PBC   & 276  &  18  &  11  &   7  &  3 \\
Hepatitis  & 155  &  19  &  6  &   13  &  2 \\
Heart Disease  & 303  &  13  &  6  &   7  &  2 \\
Heart Stat & 270  &  13  &  7  &   6  &  2 \\
Horse  & 368  &  22  &  15  &   7  &  2 \\
Autos  & 205  &  25  &  15  &   10  &  6 \\
Arrhythmia  & 452  &  279  &  206  &   73  &  2 \\
Ionosphere  & 351  &  34  &  15  &   7  &  2 \\
Credit Approval   & 690  &  15  &  6  &   9  &  2 \\
German  & 1000  &  20  &  13  &   7  &  2 \\
Contraceptive  & 1473  &  9  &  2  &   7  &  3 \\
Libras  & 360  &  90  &  90  &   0  &  15 \\
Soybean   & 307  &  35  &  0  &   35  &  4 \\ \hline
\end{tabular}}
\end{table}
\subsection{Compared methods}
We compared the proposed SFSDFC framework with five state-of-the-art mixed-type feature selection methods. MRMR \cite{peng2005feature} and CFS \cite{hall1999correlation} are selected as representatives for feature selection methods that consider both feature redundancy and relevance. Relief \cite{kononenko1994estimating} and RFS \cite{nie2010efficient} are two well-known supervised selection methods that only explore the feature relevance. The method in \cite{zhang2016feature}, termed Zhang et al., is selected from the fuzzy-rough set based feature selection approaches. 
\begin{table*}[t]
\scriptsize
\centering
\caption{Performance of SFSDFC and five state-of-the-art approaches on $k$nn in terms of accuracy. (Relative ranks are presented in parenthesis)}\label{table:2} 
\begin{tabular}{lcccccccc}
\hline
Datasets        & MRMR        & CFS         & ReliF       & RFS         & Zhang et al. & SFSDFC    & Full    & No. of Selected F \\ \hline
PBC             & 43.99\% (3) & 38.48\% (4) & 36.38\% (5) & 35.60\% (6) & \textbf{48.09\% (1)} & 47.55\% (2) & 47.92\% & 11       \\
Hepatitis        & 53.47\% (6) & 53.68\% (5) & 54.47\% (4) & 61.57\% (2) & 60.67\% (3) & \textbf{61.74\% (1)} & 62.84\% & 12       \\
Heart Diease    & 48.77\% (5) & 54.53\% (3) & 47.48\% (6) & 52.52\% (4) & 57.33\% (2) & \textbf{57.55\% (1)} & 61.98\% & 5        \\
Heart Stat      & 67.42\% (5) & 69.63\% (4) & 63.33\% (6) & \textbf{83.33\% (1)} & 81.48\% (2) & 80.14\% (3) & 81.85\% & 6        \\
Horse           & 67.02\% (6) & 70.68\% (4) & 68.56\% (5) & \textbf{72.65\% (1)} & 71.13\% (3) & 71.78\% (2) & 70.72\% & 14       \\
Autos           & 47.14\% (6) & 56.57\% (5) & 66.08\% (4) & 71.71\% (3) & \textbf{75.67\% (1)} & 72.97\% (2) & 72.31\% & 10       \\
Arrhythmia         & 63.09\% (3) & 63.33\% (2) & 62.85\% (4) & 61.67\% (5) & 57.14\% (6) & \textbf{63.81\% (1)} & 62.61\% & 53      \\
Ionosphere      & 71.51\% (2) & 69.79\% (4) & 69.50\% (5) & 68.94\% (6) & \textbf{74.29\% (1)} & 70.65\% (3) & 70.07\% & 15       \\
Credit Approval & 80.22\% (2) & 71.94\% (5) & 70.55\% (6) & 78.23\% (3) & \textbf{83.12\% (1)} & 75.46\% (4) & 84.27\% & 6        \\
German          & 68.34\% (4) & 70.50\% (2) & 68.10\% (5) & 59.40\% (6) & 68.40\% (3) & \textbf{70.60\% (1)} & 70.28\% & 8        \\
Contraceptive      & 46.57\% (4) & 45.68\% (5) & 46.64\% (3) & 42.97\% (6) & 48.50\% (2) & \textbf{49.62\% (1)} & 44.67\% & 2        \\
Libras          & 73.33\% (4) & 72.50\% (5) & 71.95\% (6) & 74.89\% (3) & 77.45\% (2) & \textbf{77.89\% (1)} & 76.78\% & 12       \\
Soybean            & 96.12\% (6) & 96.85\% (4) & 96.44\% (5) & 97.05\% (3) & 97.47\% (2) & \textbf{97.58\% (1)} & 95.64\% & 12        \\ \hline
Mean Ranks      & 4.31        & 4.00        & 4.92        & 3.77        & 2.23        & \textbf{1.77}        &         &        \\\hline 
\end{tabular}
\end{table*}
\subsection{Experimental settings}
With the python code implementation from \cite{li2018feature}, we conducted experiments on MRMR, CFS, Relief, and RFS using the default parameter settings. For the Zhang et al. method, we obtained the MATLAB code from the authors and used the settings recommended in \cite{zhang2016feature}. For all baseline methods, the size of the final selected feature subset is determined based on the number of features selected by SFDFC. The $k$NN \cite{altman1992introduction} and linear support vector machine (SVM) \cite{cortes1995support} are used as the base classifiers for performance evaluation. The value of $k$ is set to be $3$. We performed the five-fold cross-validations and repeated the experiments ten times. The average values of the evaluation metric are reported in Tables \ref{table:2} and \ref{table:3}. Tables \ref{table:2} and \ref{table:3} also summarizes the classification accuracy values with the full feature set. All experiments are conducted on an Intel Xeon (R) machine with 64GB RAM operating on Microsoft Windows 10.
\subsection{Evaluation metrics}
We utilized the classification accuracy \cite{metz1978basic} as the evaluation metric. Let $n_{i}$ be the number of samples that are correctly classified to class $i$ and $c$ be the number of classes. With a total number of $n$ samples, the classification accuracy can be expressed as follows.
\begin{equation}
	Acc=\frac{\sum_{i=1}^{c}{n_{i}}}{n}. \label{eq:8}
\end{equation}
\indent To further compare SFSDFC with the state-of-the-art methods, the friedman rank test \cite{friedman1940comparison} and nemenyi post-hoc test \cite{devijver1982pattern} are employed here to statistically analyze the difference between SFSDFC method with the other methods. 
\subsection{Result discussions}
\begin{table*}
\caption{Performance of SFSDFC and five state-of-the-art approaches on linear SVM in terms of accuracy. (Relative ranks are presented in parenthesis)} \label{table:3} 
\centering
\scriptsize
\begin{tabular}{lccccccccc}
\hline
Datasets        & MRMR        & CFS         & ReliF       & RFS         & Zhang et al. & SFSDFC    & Full    & No. of Selected F \\ \hline
PBC             & 41.35\% (5) & 39.83\% (6) & 43.06\% (4) & 46.58\% (3) & \textbf{50.70\% (1)} & 49.14\% (2) & 49.71\% & 11       \\
Hepatitis        & 64.95\% (6) & 66.58\% (5) & 67.97\% (4) & 69.05\% (3) & 71.11\% (2) & \textbf{71.58\% (1)} & 71.05\% & 12       \\
Heart Diease    & 55.45\% (6) & 55.94\% (5) & 56.12\% (4) & 62.34\% (3) & 63.33\% (2) & \textbf{63.94\% (1)} & 62.97\% & 5        \\
Heart Stat      & 74.44\% (6) & 78.15\% (4) & 76.66\% (5) & \textbf{82.97\% (1)} & 81.48\% (2) & 80.75\% (3) & 82.04\% & 6        \\
Horse           & 75.42\% (5) & 77.47\% (4) & 74.21\% (6) & \textbf{80.07\% (1)} & 79.29\% (2) & 78.53\% (3) & 80.15\% & 14       \\
Autos           & 55.34\% (6) & 57.59\% (5) & 58.43\% (4) & 60.86\% (3) & 63.25\% (2) & \textbf{63.48\% (1)} & 67.34\% & 10       \\
Arrhythmia         & 65.55\% (4) & 66.48\% (3) & 64.28\% (5) & 63.81\% (6) & 66.90\% (2) & \textbf{68.62\% (1)} & 67.23\% & 53      \\
Ionosphere      & 66.66\% (6) & 66.91\% (5) & 67.22\% (4) & 68.35\% (3) & \textbf{69.14\% (1)} & 68.37\% (2) & 69.82\% & 15       \\
Credit Approval & 76.67\% (3) & 71.96\% (6) & 74.65\% (5) & 78.34\% (2) & \textbf{78.51\% (1)} & 73.16\% (4) & 76.50\% & 6        \\
German          & 74.30\% (5) & 73.91\% (6) & 74.61\% (4) & 75.19\% (3) & 75.80\% (2) & 75.88\% (1) & 75.65\% & 8        \\
Contraceptive      & 46.98\% (4) & 47.18\% (3) & 42.83\% (5) & 42.02\% (6) & \textbf{49.32\% (1)} & 48.27\% (2) & 47.57\% & 2        \\
Libras          & 56.38\% (3) & 54.12\% (4) & 51.25\% (6) & 51.94\% (5) & 60.12\% (2) & \textbf{60.28\% (1)} & 70.27\% & 12       \\
Soybean            & 96.87\% (3) & 96.45\% (4) & 95.85\% (5) & 95.68\% (6) & 97.12\% (2) & \textbf{97.79\% (1)} & 96.41\% & 12        \\ \hline
Mean Ranks      & 4.85        & 4.77        & 4.69        & 3.46        & \textbf{1.69}        & \textbf{1.77}        &         &             \\ \hline
\end{tabular}
\end{table*}
\begin{figure}
\centering
\includegraphics[width=55mm]{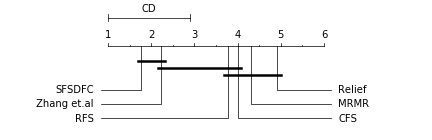}
\caption{Comparison of SFSDFC against other baseline methods with the Nemenyi test with $\alpha=0.05$ in terms of $k$NN.}
\label{fig: cd diagrams1}
\end{figure}
Table \ref{table:2} presents the comparison results of SFSDFC with the state-of-the-art approaches evaluated using a $k$NN classifier. From Table \ref{table:2}, the proposed supervised feature selection method outperforms other baseline methods on the majority of the datasets and achieves the highest average rank. Specifically, SFSDFC shows better accuracy on Hepatitis, Heart Disease, Arrhythmia, German, Contraceptive, Libras, and WPBC datasets. For the remaining datasets, although SFSDFC provides slightly lower accuracy values than Zhang et al. and RFS methods, Fig. \ref{fig: cd diagrams1} indicates that SFSDFC is statistically comparable with Zhang et al. method. With a statistically comparable performance, SFSDFC is capable of discovering an appropriate final feature subset without the sequential feature selection procedure. 
\begin{figure}
\centering
\includegraphics[width=55mm]{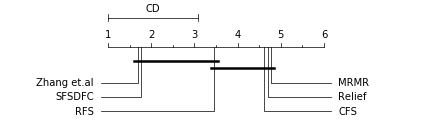}
\caption{Comparison of SFSDFC against other methods with the Nemenyi test with $\alpha=0.05$ in terms of SVM.}
\label{fig: cd diagrams2}
\end{figure}\\
\indent The comparison results between SFSDFC and other approaches evaluated with a linear SVM are shown in Table \ref{table:3}. Table \ref{table:3} demonstrates that SFSDFC shows better performance than all compared methods on seven benchmark datasets. In terms of average ranks, Zhang et al. achieves the highest average rank of $1.69$. According to the statistical analysis from Fig. \ref{fig: cd diagrams2}, SFSDFC is statistically comparable with RFS and Zhang et al. methods while it presents statistically better performance than MRMR, CFS, and Relief approaches. 

\subsection{\textcolor{black}{Time complexity comparison}}
From the experimental results, it is observed that SFSDFC achieves statistically comparable performance with Zhang et al. method on both $k$NN and linear SVM classifiers. To further compare their performance, we present the time complexity comparison in Table \ref{table:complexity}. 
\begin{table}[h]
\centering
\scriptsize
\caption{Time complexity comparison.} \label{table:complexity}
\begin{tabular}{l|c}
\hline
Compared Methods      & Time Complexity \\ \hline
Zhang et al. \cite{zhang2016feature} &       $O(n^2m^2)$          \\
Proposed       &       $O(nm^2)$         \\ \hline
\end{tabular}
\end{table}\\
\indent \textcolor{black}{As shown in Table \ref{table:complexity}, SFSDFC requires less time complexity and does not require a sequential feature selection procedure to determine the size of the final feature subset.}
\section{CONCLUSIONS} \label{section:6}
\textcolor{black}{In this paper, a supervised mixed-data feature selection method through density-based feature clustering, namely SFSDFC, is presented to automatically discover an appropriate feature subset. SFSDFC separated features into continuous \& discrete feature subsets, and then performed the feature selection independently. We employed a novel density-based feature clustering procedure to partition the feature space into an appropriate number of feature clusters with no predefined parameters. A simple yet effective feature selection strategy is introduced for SFSDFC to select both relevant and representative features from feature clusters. Empirical results and comparison studies proved that SFSDFC always provided statistically comparable or better performance than the existing approaches. The complexity analysis indicates that the time complexity of SFSDFC increases linearly as the sample size increases while it has a quadratic relationship with feature size.} 

In the future, we will focus on reducing the time complexity of the SFSDFC method.
Moreover, efforts will be made to investigate effective strategies for feature type identification.

\section*{ACKNOWLEDGMENT}
This work is supported by the Air Force Research Laboratory and the OSD under agreement number FA8750-15-2-0116. Also, this work is partially funded by the NASA University Leadership Initiative (ULI), the National Science Foundation, National Institute of Aerospace's Langley Distinguished Professor Program, and the OSD RTL under grants number 80NSSC20M0161, 2000320, C16-2B00-NCAT, and W911NF-20-2-0261 respectively. The authors would like to thank them for their support.



\bibliographystyle{IEEEtran}
\bibliography{refers}
\addtolength{\textheight}{-12cm}

\end{document}